\documentclass[11pt]{article}
\usepackage{colacl}
\usepackage{epsfig}
\usepackage{amsmath}
\usepackage{amssymb}

\def\pp{\par\noindent}

\title{{\LARGE A Learning Approach to Shallow Parsing}\thanks{Research 
supported by NSF grants IIS-9801638 and SBR-9873450.}}

\author{{\Large 
        Marcia Mu\~{n}oz\thanks{Research supported by NSF grant CCR-9502540.}~~~~~ 
	Vasin Punyakanok$^*$~~~~~
	Dan Roth$^*$~~~~~
	Dav Zimak}  \\
\\
	Department of Computer Science \\
	University of Illinois at Urbana-Champaign \\
	Urbana, IL 61801 USA
	}

\begin{document}
\maketitle

\begin{abstract}
{\small 
A SNoW based learning approach to shallow parsing tasks is presented
and studied experimentally.  The approach learns to identify syntactic
patterns by combining simple predictors to produce a coherent
inference.  Two instantiations of this approach are studied and
experimental results for Noun-Phrases (NP) and Subject-Verb (SV)
phrases that compare favorably with the best published results are
presented.
In doing that, we compare two ways of modeling the problem of
learning to recognize patterns and suggest that shallow parsing
patterns are better learned using open/close predictors than
using inside/outside predictors.}
\end{abstract}

\section{Introduction}
Shallow parsing is studied as an alternative to full-sentence parsers.  Rather
than producing a complete analysis of sentences, the alternative is to perform
only partial analysis of the syntactic structures in a
text~\cite{Harris57,Abney91,Greffenstette93}.
Shallow parsing information such as NPs and other syntactic sequences have
been found useful in many large-scale language processing applications
including information extraction and text summarization.
A lot of the work on shallow parsing over the past years has
concentrated on manual construction of rules. The observation that
shallow syntactic information can be extracted using local information
-- by examining the pattern itself, its nearby context and the local
part-of-speech information -- has motivated the use of learning
methods to recognize these
patterns~\cite{Church88,RamshawMa95,ArgamonDaKr98,CardiePi98}.

This paper presents a general learning approach for identifying syntactic
patterns, based on the SNoW learning architecture~\cite{Roth98,Rothtech99}.
The SNoW learning architecture is a sparse network of linear functions over a
pre-defined or incrementally learned feature space.
SNoW is specifically tailored for learning in domains in which the potential
number of information sources (features) taking part in decisions is very
large -- of which NLP is a principal example.  Preliminary versions of it have
already been used successfully on several tasks in natural
language processing~\cite{Roth98,GoldingRo99,RothZe98}.
In particular, SNoW's sparse architecture supports well chaining and
combining predictors to produce a coherent inference. This property of the
architecture is the base for the learning approach studied here in the context
of shallow parsing.

Shallow parsing tasks often involve the identification of syntactic phrases or
of words that participate in a syntactic relationship.  Computationally, each
decision of this sort involves multiple predictions that interact in some way.
For example, in identifying a phrase, one can identify the beginning and
end of the phrase while also making sure they are coherent.

Our computational paradigm suggests using a SNoW based predictor as a building
block that learns to perform each of the required predictions, and writing a
simple program that activates these predictors with the appropriate input,
aggregates their output and controls the interaction between the predictors.
Two instantiations of this paradigm are studied and evaluated on two different
shallow parsing tasks -- identifying base NPs and SV phrases.
The first instantiation of this paradigm uses predictors to decide whether
each word belongs to the interior of a phrase or not, and then groups the
words into phrases. The second instantiation finds the borders of phrases
(beginning and end) and then pairs them in an ``optimal'' way into different
phrases. These problems formulations are similar to those studied
in~\cite{RamshawMa95} and~\cite{Church88,ArgamonDaKr98}, respectively.

The experimental results presented using the SNoW based approach compare
favorably with previously published results, both for NPs and SV phrases.  As
important, we present a few experiments that shed light on some of the issues
involved in using learned predictors that interact to produce the desired
inference. In particular, we exhibit the contribution of \emph{chaining}:
features that are generated as the output of one of the predictors contribute
to the performance of another predictor that uses them as its input.
Also, the comparison between the two instantiations of the learning paradigm
-- the Inside/Outside and the Open/Close -- shows the advantages of the
Open/Close model over the Inside/Outside, especially for the task of
identifying long sequences.

The contribution of this work is in improving the state of the art in
learning to perform shallow parsing tasks, developing a better understanding
for how to model these tasks as learning problems and in further studying the
SNoW based computational paradigm that, we believe, can be used in many other
related tasks in NLP.

The rest of this paper is organized as follows: The SNoW architecture is
presented in Sec.~\ref{sec:snow}.  Sec.~\ref{sec:modeling} presents the
shallow parsing tasks studied and provides details on
the computational approach.  Sec.~\ref{sec:methodology} describes the data
used and the experimental approach, and Sec.~\ref{sec:results} presents and
discusses the experimental results.

\section{SNoW}
\label{sec:snow}

The SNoW (Sparse Network of Winnows\footnote{To winnow: to separate chaff from
grain.}) learning architecture is a sparse network of linear units over a
common pre-defined or incrementally learned feature space.  
Nodes in the input layer of the network represent simple relations
over the input sentence and are being used as the input features. Each
linear unit is called a \emph{target node} and represents relations
which are of interest over the input sentence; in the current
application, target nodes may represent a potential prediction with
respect to a word in the input sentence, e.g., \emph{inside a phrase},
\emph{outside a phrase}, \emph{at the beginning of a phrase}, etc.
An input sentence, along with a designated word of interest in it, is mapped
into a set of features which are active in it; this representation is
presented to the input layer of SNoW and propagates to the target nodes.
Target nodes are linked via weighted edges to (some of the) input features.
Let $\mathcal{A}_{t} = \{i_1, \ldots, i_m \}$ be the set of features that are
active in an example and are linked to the target node $t$. Then the linear
unit is \emph{active} iff $\sum_{i \in \mathcal{A}_{t}} w^t_i > \theta_t$,
where $w^t_i$ is the weight on the edge connecting the $i$th feature to the
target node $t$, and $\theta_t$ is the threshold for the target node $t$.

Each SNoW \emph{unit} may include a collection of subnetworks, one for
each of the target relations.
A given example is treated autonomously by each target subnetwork; an
example labeled $t$ may be treated as a positive example by the subnetwork
for $t$ and as a negative example by the rest of the target nodes.

The learning policy is on-line and mistake-driven; several update rules can be
used within SNoW. The most successful update rule, and the only one used in
this work is a variant of Littlestone's~\shortcite{Littlestone88} Winnow
update rule, a multiplicative update rule tailored to the situation in which
the set of input features is not known a priori, as in the infinite attribute
model~\cite{Blum92}.
This mechanism is implemented via the sparse architecture of SNoW. That is,
(1) input features are allocated in a data driven way -- an input node for the
feature $i$ is allocated only if the feature $i$ was active in any input
sentence and (2) a link (i.e., a non-zero weight) exists between a target node
$t$ and a feature $i$ if and only if $i$ was active in an example labeled $t$.

The Winnow update rule has, in addition to the threshold $\theta_t$ at
the target $t$, two update parameters: a {\it promotion\/} parameter
$\alpha > 1$ and a {\it demotion\/} parameter $0 < \beta < 1$.  These
are being used to update the current representation of the target $t$
(the set of weights $w_i^t$) only when a mistake in prediction is
made.
Let ${\cal A}_t = \{i_1, \ldots, i_m \}$ be the set of active features
that are linked to the target node $t$.
If the algorithm predicts $0$ (that is, $\sum_{i \in \mathcal{A}_{t}}
w^t_i \leq \theta_t$) and the received label is $1$, the active weights
in the current example are {\em promoted} in a multiplicative fashion:
$\forall i \in {\cal A}_t, w^t_i \leftarrow \alpha \cdot w^t_i.$
If the algorithm predicts $1$ ($\sum_{i \in \mathcal{A}_{t}} w^t_i >
\theta_t$) and the received label is $0$, the active weights in the
current example are {\em demoted}\: $\forall i \in {\cal A}_t, w^t_i
\leftarrow \beta \cdot w^t_i.$ All other weights are unchanged.

The key feature of the Winnow update rule is that the number of
examples required to learn a linear function grows linearly with
the number of \emph{relevant\/} features and only logarithmically with
the total number of features. This property seems crucial in domains
in which the number of potential features is vast, but a relatively
small number of them is relevant.  Winnow is known to learn
efficiently any linear threshold function and to be robust in the
presence of various kinds of noise and in cases where no
linear-threshold function can make perfect classifications, while
still maintaining its abovementioned dependence on the number of total
and relevant attributes~\cite{Littlestone91,KivinenWa95}.

Once target subnetworks have been learned and the network is being evaluated,
a decision support mechanism is employed, which selects the dominant active
target node in the SNoW unit via a winner-take-all mechanism to produce a
final prediction. The decision support mechanism may also be cached and
processed along with the output of other SNoW units to produce a coherent
output.

\begin{figure*}[!htb]
\begin{center}
\epsfig{file=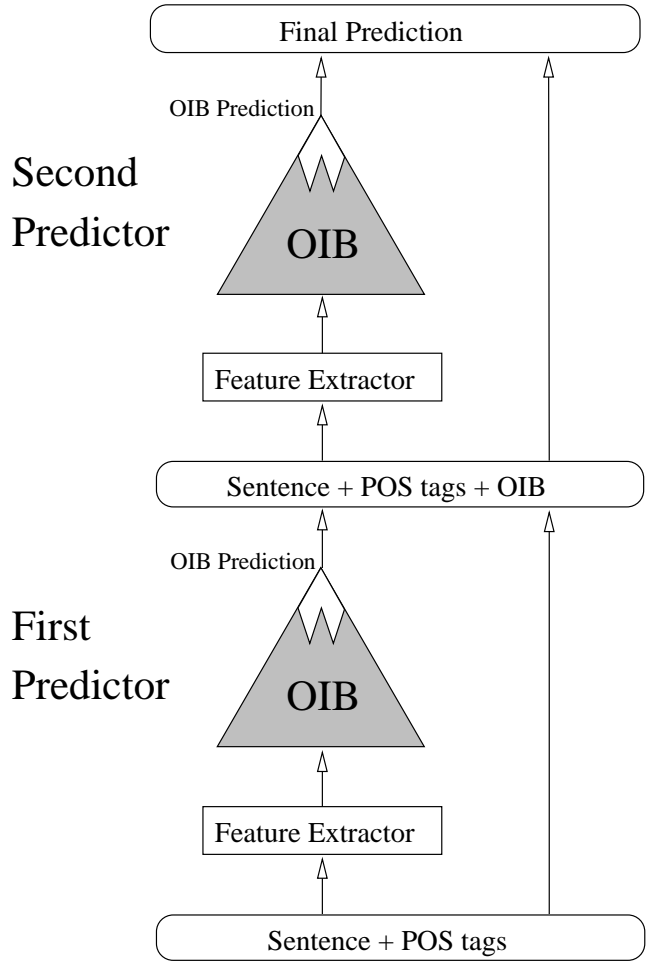}
\caption{Architecture for Inside/Outside method.}
\label{fig:OIBArch}
\end{center}
\end{figure*}
\begin{figure*}[!htb]
\begin{center}
\epsfig{file=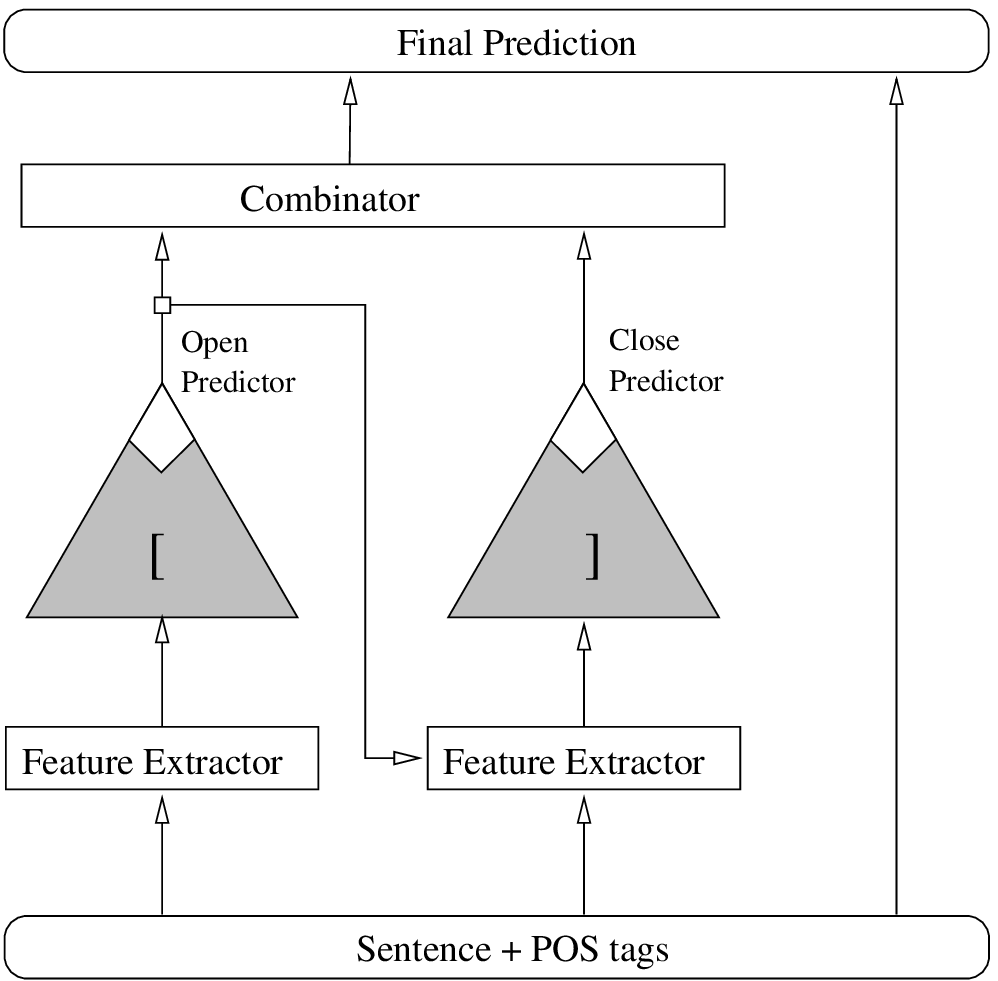}
\caption{Architecture for Open/Close Method.}
\label{fig:OpenCloseArch}
\end{center}
\end{figure*}

\section{Modeling Shallow Parsing}
\label{sec:modeling}

\subsection{Task Definition}

This section describes how we model the shallow parsing tasks studied here as
learning problems. The goal is to detect NPs and SV phrases.  Of the several
slightly different definitions of a base NP in the literature we use for the
purposes of this work the definition presented in~\cite{RamshawMa95} and used
also by~\cite{ArgamonDaKr98} and others. That is, a base NP is a non-recursive
NP that includes determiners but excludes post-modifying prepositional phrases
or clauses.  For example:

\pp
{\tt
...presented 
\:[\:last year\:]\: 
in 
\:[\:Illinois\:]\:
in front of ...
}

SV phrases, following the definition suggested in~\cite{ArgamonDaKr98}, are
word phrases starting with the subject of the sentence and ending with the
first verb, excluding modal verbs\footnote{Notice that according
to this definition the identified verb may not correspond to the subject, but
this phrase still contains meaningful information; in any case, the learning
method presented is independent of the specific definition used.}.
For example, the SV phrases are bracketed in the following:

\pp
{\tt
...presented 
\:[\:a theory that claims\:]\:
that 
\:[\:the algorithm runs\:]\:
and performs...
}

Both tasks can be viewed as sequence recognition problems. This can be modeled
as a collection of prediction problems that interact in a specific way.  For
example, one may predict the first and last word in a target sequence.
Moreover, it seems plausible that information produced by one predictor (e.g.,
predicting the beginning of the sequence) may contribute to others (e.g.,
predicting the end of the sequence).

Therefore, our computational paradigm suggests using SNoW predictors that
learn separately to perform each of the basic predictions, and \emph{chaining}
the resulting predictors at evaluation time.  Chaining here means that the
predictions produced by one of the predictors may be used as (a part of the)
input to others\footnote{The input data used in all the experiments presented
here consists of part-of-speech tagged data. In the demo of the system
(available from {\tt http://l2r.cs.uiuc.edu/\~{}cogcomp/eoh/index.html}),
an additional layer of chaining is used. Raw sentences are supplied as input
and are processed using a SNoW based POS tagger~\cite{RothZe98} first.}.

Two instantiations of this paradigm -- each of which models the problems using
a different set of predictors -- are described below.

\subsection{Inside/Outside Predictors}
The predictors in this case are used to decide, for each word, whether
it belongs to the interior of a phrase or not; this information is then used
to group the words into phrases. Since annotating words only with
Inside/Outside information is ambiguous in cases of two consecutive phrases,
an additional predictor is used.  Specifically, each word in the sentence may
be annotated using one of the following labels: {\bf O} - the current word is
outside the pattern. {\bf I} - the current word is inside the pattern.  {\bf
B} - the current word marks the beginning of a pattern that immediately
follows another pattern\footnote{There are other ways to define the B
annotation, e.g., as always marking the beginning of a phrase. The modeling
used, however, turns out best experimentally.}.

For example, the sentence {\tt I went to California last May} would be marked
for base NPs as: 

\begin{center}
\begin{tabular}{cccccc}
{\tt I} & {\tt went} & {\tt to} & {\tt California} & {\tt last} & {\tt May} \\
I &  O   &  O &    I       &  B   &  I  \\
\end{tabular}
\end{center}
indicating that the NPs are {\tt I}, {\tt California} and {\tt last May}.
This approach has been studied in~\cite{RamshawMa95}.

\subsubsection{Architecture}

SNoW is used in order to learn the {\bf OIB} annotations both for NPs and SV
phrases. In each case, two predictors are learned, which differ in the type of
information they receive in their input.
A \emph{first predictor} takes as input a sentence along with the corresponding
part-of-speech (POS) tags. The features extracted from this input represent
the local context of each word in terms of POS tags (with the possible
addition of lexical information), as described in Sec~\ref{subsec:features}.
The SNoW predictor in this case consists of three targets -- \textbf{O},
\textbf{I} and \textbf{B}.  Figure~\ref{fig:OIBArch} depicts the feature
extraction module which extracts the local features and generates an example
for each word in the sentence. Each example is labeled with one of \textbf{O},
\textbf{I} or \textbf{B}.

The \emph{second predictor} takes as input a sentence along with the
corresponding POS tags as well as the {\bf Inside/Outside information}. The
hope is that representing the local context of a word using the Inside/Outside
information for its neighboring words, in addition to the POS and lexical
information, will enhance the performance of the predictor.  While this
information is available during training, since the data is annotated with the
{\bf OIB} information, it is not available in the input sentence at evaluation
time.
Therefore, at evaluation time, given a sentence (represented as a sequence of
POS tags), we first need to evaluate the first predictor on it, generate an
Inside/Outside representation of the sentence, and then use this to generate
new features that feed into the second predictor. 

\begin{figure*}[!htb]
\begin{center}
\begin{tabular}{cccccccccccccc}

            & $s_1$ &             & $s_2$ &             & $s_3$ &            & $s_4$ &             & $s_5$ &
            & $w_6$ & \\ \hline\hline
$[_{0.8}^*$ &       & $]_{0.3}^*$ &       &             &       &            &       & $]_{0.6}$   &       &
            &       &            \\
            &       &             &       & $[_{0.5}^*$ &       &  $]_{0.3}$ &       & $]_{0.9}^*$ &       &
 $]_{0.3}$  &       &            \\
            &       &             &       &             &       &  $[_{0.4}$ &       &   $]_{0.5}$ &       &
            &       &  $]_{0.3}$   \\ \hline
[\:\:\:\:   & $s_1$ &  ]\:\:\:\: & $s_2$ &   [\:\:\:\: & $s_3$ &            & $s_4$ &   ]\:\:\:\: & $s_5$ &
            & $s_6$ & \\

\end{tabular}
\caption{Example of combinator assignment.  
Subscripts denote the confidence of the bracket candidates.  Bracket
candidates that would be chosen by the combinator are marked with a
*.}
\label{fig:combinator}
\end{center}
\end{figure*}

\subsection{Open/Close Predictors}	
The predictors in this case are used to decide, for each word, whether it is
the first in a phrase, the last in a phrase, both of these, or none of these.
In this way, the phrase boundaries are determined; this is annotated by
placing an open bracket (\:[\:) before the first word and a close bracket
(\:]\:) after the last word of each phrase.
Our earlier example would be marked for base NPs as: \texttt{[I] went to
[California] [last May]}.
This approach has been studied in~\cite{Church88,ArgamonDaKr98}.

\subsubsection{Architecture}

The architecture used for the Open/Close predictors is shown in
Figure~\ref{fig:OpenCloseArch}.  Two SNoW predictors are used, one to predict
if the word currently in consideration is the first in the phrase (an open
bracket), and the other to predict if it is the last (a close bracket).
Each of the two predictors is a SNoW network with two competing target nodes:
one predicts if the current position is an open (close) bracket and the other
predicts if it is not.
In this case, the actual activation value (sum of weights of the
active features for a given target) of the SNoW predictors is used to
compute a \emph{confidence} in the prediction.  Let $t_Y$ be the
activation value for the {\em yes-bracket} target and $t_N$ for the {\em
no-bracket} target. Normally, the network would predict the target
corresponding to the higher activation value.  In this case, we prefer
to cache the system preferences for each of the open (close) brackets
predictors so that several bracket pairings can be considered when all
the information is available.
The confidence, $\gamma$, of a candidate is defined by $\gamma = t_Y /
(t_Y + t_N)$.  Normally, SNoW will predict that there is a bracket if
$\gamma \geqslant 0.5$, but this system employs an threshold $\tau$.
We will consider any bracket that has $\gamma \geqslant \tau$ as a
candidate.  The lower $\tau$ is, the more candidates will be
considered.

The input to the open bracket predictor is a sentence and the POS tags
associated with each word in the sentence.  For each position in the
sentence, the open bracket predictor decides if it is a candidate for
an open bracket.
For each open bracket candidate, features that correspond to this
information are generated; the close bracket predictor can
(potentially) receive this information in addition to the sentence and
the POS information, and use it in its decision on whether a given
position in the sentence is to be a candidate for a close bracket
predictor (to be paired with the open bracket candidate).

\begin{figure*}[!htb]
\begin{center}
\begin{tabular}{cccccccccc}
     &     &    &         &     & $\Downarrow$ & & & & \\
This & is  & an & example &  of & how & to & generate & features & . \\
DT   & VBZ & DT & NN      & IN  & WRB & TO & VB       &    NNS   & . \\ 
\\
\end{tabular}

\begin{tabular}{lllll}
\hline
Conj. Size  & \multicolumn{1}{c}{1} &  \multicolumn{1}{c}{2} &  \multicolumn{1}{c}{3} &
		 \multicolumn{1}{c}{4} \\ \hline
For Tags    & DT \_ \_ ()   & DT NN \_ ()  & DT NN IN ()  & DT NN IN (WRB)  \\ 
$w=3$       & NN \_ ()      & NN IN ()     & NN IN (WRB)  & NN IN (WRB) TO  \\
$k=4$       & IN ()         & IN (WRB)     & IN (WRB) TO  & IN (WRB) TO VB  \\
            & (WRB)         & (WRB) TO     & (WRB) TO VB  & (WRB) TO VB NNS \\
            & () TO         & () TO VB     & () TO VB NNS & \\
            & () \_ VB      & () \_ VB NNS &              & \\
            &  () \_ \_ NNS & & & \\  \hline
For Words   & of ()         & of (how) \\
$w=1$       & (how)         & (how) to \\
$k=2$       & () to \\ \hline
\end{tabular}
\end{center}
\caption{An example of feature extraction.}
\label{FeatEx}
\end{figure*}

\begin{table*}[!htb]
\vspace*{0.2in}
\begin{center}
\begin{tabular}{|l|r|r|r|}
\hline
Data     & Sentences & Words & NP Patterns \\
\hline\hline
Training & 8936      & 211727 & 54758 \\
Test     & 2012      &  47377 & 12335 \\
\hline
\end{tabular}
\caption{Sizes of the training and test data sets for NP Patterns.}
\label{NPSizeTable}
\end{center}
\end{table*}

\begin{table*}[!htb]
\vspace*{0.2in}
\begin{center}
\begin{tabular}{|l|r|r|r|}
\hline
Data     & Sentences & Words & SV Patterns \\
\hline\hline
Training & 16397    & 394854 & 25024 \\
Test     & 1921     &  46451 & 3044 \\
\hline
\end{tabular}
\caption{Sizes of the training and test data sets for SV Patterns.}
\label{SVSizeTable}
\end{center}
\end{table*}

\subsubsection{Combinator}

Finding the final phrases by pairing the open and close bracket
candidates is crucial to the performance of the system; even given
good prediction performance choosing an inadequate pairing would
severely lower the overall performance.  We use a graph based method
that uses the confidence of the SNoW predictors to generate the
consistent pairings, at only a linear time complexity.

We call $p=(o,c)$ a \emph{pair}, where $o$ is an open bracket and
$c$ is any close bracket that was predicted with respect to $o$. 
The \emph{position} of a bracket at the $i$th word is defined to be
$i$ if it is an open bracket and $i+1$ if it is a close bracket.
Clearly, a pair $(o, c)$ is possible only when $pos(o) < pos(c)$.
The confidence of a bracket $t$ is the weight $\gamma(t)$. The
value of a pair $p = (o, c)$ is defined to be $v(p) = \gamma(o)*\gamma(c)$.  
The pair $p_1$ \emph{occurs before} the pair $p_2$ if $pos(c_1) \leqslant
pos(o_2)$.
$p_1$ and $p_2$ are \emph{compatible} if either $p_1$ occurs before
$p_2$ or $p_2$ occurs before $p_1$.  A \emph{pairing} is a set of
pairs $P = \{p_1, p_2, \ldots p_n\}$ such that $p_i$ is compatible
with $p_j$ for all $i$ and $j$ where $i \not= j$.  The value of the
pairing is the sum of all of the values of the pairs within the
pairing.

Our combinator finds the pairing with the maximum value.  Note that
while there may be exponentially many pairings, by modeling the
problem of finding the maximum valued pairing as a shortest path
problem on a directed acyclic graph, we provide a linear time solution.
Figure~\ref{fig:combinator} gives an example of pairing bracket
candidates of the sentence $S = s_1 s_2 s_3 s_4 s_5 s_6$, where the
confidence of each candidate is written in the subscript.

\subsection{Features}
\label{subsec:features}

The features used in our system are relational features over the
sentence and the POS information, which can be defined by a pair of
numbers, $k$ and $w$.  Specifically, features are either word
conjunctions or POS tags conjunctions. All conjunctions of size up to
$k$ and within a symmetric window that includes the $w$ words before and after
the designated word are generated.

An example is shown in Figure~\ref{FeatEx} where $(w, k) = (3, 4)$ for POS
tags, and $(w, k) = (1, 2)$ for words.  In this example the word ``how'' is
the designated word with POS tag ``WRB''.  ``()'' marks the position of the
current word (tag) if it is not part of the feature, and ``(how)'' or
``(WRB)'' marks the position of the current word (tag) if it is part of the
current feature.  The distance of a conjunction from the current word (tag)
can be induced by the placement of the special character ``\_'' in the
feature.  We do not consider mixed features between words and POS tags as in
\cite{RamshawMa95}, that is, a single feature consists of \emph{either}
words \emph{or} tags.

Additionally, in the Inside/Outside model, the second predictor incorporates
as features the OIB status of the $w$ words before and after the designated
word, and the conjunctions of size 2 of the words surrounding it.


\section{Methodology}
\label{sec:methodology}

\subsection{Data}

In order to be able to compare our results with the results obtained by other
researchers, we worked with the same data sets already used
by~\cite{RamshawMa95,ArgamonDaKr98} for NP and SV detection. These data sets
were based on the Wall Street Journal corpus in the Penn
Treebank~\cite{wsj-corpus}.  For NP, the training and test corpus was prepared
from sections 15 to 18 and section 20, respectively; the SV corpus was
prepared from sections 1 to 9 for training and section 0 for testing.
Instead of using the NP bracketing information present in the tagged Treebank
data, Ramshaw and Marcus modified the data so as to include bracketing
information related only to the non-recursive, base NPs present in each
sentence while the subject verb phrases were taken as is. The data sets
include POS tag information generated by Ramshaw and Marcus using Brill's
transformational part-of-speech tagger~\cite{Brill95}.

The sizes of the training and test data are summarized in
Table~\ref{NPSizeTable} and Table~\ref{SVSizeTable}.

\subsection{Parameters}

The Open/Close system has two adjustable parameters, $\tau_{\:[}$ and
$\tau_{\:]}$, the threshold for the open and close bracket predictors,
respectively.  For all experiments, the system is first trained on 90\% of the
training data and then tested on the remaining 10\%. The $\tau_{\:]}$ and
$\tau_{\:[}$ that provide the best performance are used on the real test
file.  After the best parameters are found, the system is trained on the
whole training data set.  Results are reported in terms of recall, precision,
and $F_{\beta}$.  $F_{\beta}$ is always used as the single value to compare
the performance.

For all the experiments, we use 1 as the initial weight, 5 as the threshold,
1.5 as $\alpha$, and 0.7 as $\beta$ to train SNoW, and it is always trained
for 2 cycles.

\subsection{Evaluation Technique}
To evaluate the results, we use the following metrics:

\[ \mbox{Recall} = \frac{\mbox{Number of correct proposed patterns}}
	{\mbox{Number of correct patterns}} \]
\[ \mbox{Precision} = \frac{\mbox{Number of correct proposed patterns} }
	{\mbox{Number of proposed patterns}} \]
\[ F_{\beta} = \frac{(\beta^{2} + 1) \cdot \mbox{Recall} \cdot \mbox{Precision}}
	{\beta^{2} \cdot \mbox{Precision} + \mbox{Recall}} \]
\[ \mbox{Accuracy} = \frac{\mbox{Number of words labeled correctly}}
	{\mbox{Total number of words}} \]

We use $\beta=1$.
Note that, for the Open/Close system, we must measure the accuracy for the open
predictor and the close predictor separately since each word can be labeled as
``Open'' or ``Not Open'' and, at the same time, ``Close'' or ``Not Close''.

\begin{table*}[!htb]
\begin{center}
\begin{tabular}{|l|c|c|c|c|}
\hline
\multicolumn{1}{|l}{Method} & \multicolumn{1}{|c}{Recall} &
\multicolumn{1}{|c}{Precision} & \multicolumn{1}{|c}{$F_{\beta=1}$} &
\multicolumn{1}{|c|}{Accuracy} \\
\hline
\hline
First Predictor                              & 90.5   & 89.8   & 90.1   & 96.9   \\
Second Predictor                             & 90.5   & 90.4   & 90.4   & 97.0   \\
First Predictor + lexical              & 92.5   & 92.2   & 92.4   & 97.6   \\
Second Predictor + lexical             & 92.5   & 92.1   & 92.3   & 97.6   \\
\hline
\end{tabular}
\caption{Results for NP detection using Inside/Outside method.}
\label{OIRresults}
\end{center}
\end{table*}
\begin{table*}[!htb]
\vspace*{0.2in}
\begin{center}
\begin{tabular}{|l|c|c|c|c|c|}
\hline
Method	& Recall & Precision & $F_{\beta = 1}$ & \multicolumn{2}{c|}{Accuracy} \\
\cline{5-6}
\	&	&	&	& Open	& Close\\
\hline\hline
SV w/o lexical	& 88.3 	& 87.9 	& 88.1 	& 98.6	& 99.4 \\
SV with lexical	& 91.9 	& 92.2 	& 92.0 	& 99.2  & 99.4 \\
NP w/o lexical	& 90.9 	& 90.3 	& 90.6 	& 97.4	& 97.8 \\
NP with lexical	& 93.1 	& 92.4 	& 92.8 	& 98.1	& 98.2 \\
\hline
\end{tabular}
\caption{Results for SV Phrase and NP detection using Open/Close method.}
\label{OCResults}
\end{center}
\end{table*}

\begin{table*}[!htb]
\vspace*{0.2in}
\begin{center}
\begin{tabular}{|c||c|c||c|c|}
\hline
$\tau$ & \multicolumn{2}{c||}{Without Open info}
& \multicolumn{2}{c|}{With Open bracket info} \\ \cline{2-5}
& Overall & Positive Only & Overall & Positive Only \\
\hline\hline
0.5 & 99.3 & 92.7 & 99.4 & 95.0 \\
\hline
\end{tabular}
\caption{Accuracy of close bracket predictor when using features created on
local information alone versus using additional features created from the open
bracket candidate.  Overall performance and performance on positive examples
only is shown.}
\label{OCAccuracy}
\end{center}
\end{table*}

\section{Experimental Results}
\label{sec:results}

\subsection{Inside/Outside}
The results of each of the predictors used in the Inside/Outside method
are presented in Table~\ref{OIRresults}.
The results are comparable to other results reported using the Inside/Outside
method~\cite{RamshawMa95} (see Table~\ref{CompNPResults}.
We have observed that most of the mistaken predictions of base NPs
involve predictions with respect to conjunctions, gerunds, adverbial
NPs and some punctuation marks.
As reported in~\cite{ArgamonDaKr98}, most base NPs present in the data are
less or equal than 4 words long. This implies that our predictors tend to
break up long base NPs into smaller ones.

The results also show that lexical information improves the
performance by nearly 2\%. This is similar to results in the
literature~\cite{RamshawMa95}.
What we found surprising is that the second predictor, that uses additional
information about the {\bf OIB} status of the local context, did not do much
better than the first predictor, which relies only on POS and lexical
information. A control experiment has verified that this is not due to the
noisy features that the first predictor supplies to the second predictor.

Finally, the Inside/Outside method was also tested on predicting SV phrases,
yielding poor results that are not shown here. An attempt at explaining this
phenomena by breaking down performance according to the length of the phrases
is discussed in Sec.~\ref{sec:discussion}. 

\subsection{Open/Close}
The results of the Open/Close method for NP and SV phrases are presented in
Table~\ref{OCResults}.  In addition to the good overall performance, the
results show significant improvement by incorporating the lexical information
into the features.
In addition to the recall/precision results we have also presented the
accuracy of each of the Open and Close predictors. These are important since
they determine the overall accuracy in phrase detection. It is evident that
the predictors perform very well, and that the overall performance degrades
due to inconsistent pairings.

An important question in the learning approach presented here is investigating
the gain achieved due to chaining. That is, whether the features extracted
from open brackets can improve the performance of the the close bracket
predictor.
To this effect, we measured the accuracy of the close bracket predictor
itself, on a word basis, by supplying it features generated from {\em correct}
open brackets.  We compared this with the same experiment, only this time
without incorporating the features from open brackets to the close bracket
predictor. 
The results, shown in Table~\ref{OCAccuracy} indicate a significant
contribution due to chaining the features. Notice that the overall accuracy
for the close bracket predictor is very high. This is due to the fact that, as
shown in Table~\ref{SVSizeTable}, there are many more negative examples than
positive examples. Thus, a predictor that always predicts ``no'' would have
an accuracy of 93.4\%.  Therefore, we considered also the accuracy over
positive examples, which indicates the significant role of the chaining. 

\subsection{Discussion}
\label{sec:discussion}

Both methods we study here -- Inside/Outside and Open/Close -- have
been evaluated before (using different learning methods) on similar
tasks. However, in this work we have allowed for a fair comparison
between two different models by using the same basic learning method
and the same features.

Our main conclusion is with respect to the robustness of the methods to
sequences of different lengths.  While both methods give good results for the
base NP problem, they differ significantly on the SV tasks. Furthermore, our
investigation revealed that the Inside/Outside method is very sensitive to the
length of the phrases.
Table~\ref{CompareLength} shows a breakdown of the performance of the two
methods on SV phrases of different lengths. Perhaps this was not observed
earlier since \cite{RamshawMa95} studied only base NPs, most of which are
short.  The conclusion is therefore that the Open/Close method is more robust,
especially when the target sequences are longer than a few tokens.

Finally, Tables~\ref{CompNPResults} and~\ref{CompSVResults} present
a comparison of our methods to some of the best NP and SV results
published on these tasks.

\begin{table*}[!htb]
\begin{center}
\begin{tabular}{|l|c||c|c|c||c|c|c|}
\hline
Length	& Patterns & \multicolumn{3}{|c||}{Inside/Outside}
& \multicolumn{3}{|c|}{Open/Close} \\
\cline{3-8}
	&	& Recall & Precision & $F_{\beta=1}$ & Recall & Precision
& $F_{\beta=1}$\\
\hline \hline
$\leqslant 4$	& 2212 	& 90.5	& 61.5	& 73.2	& 94.1 	& 93.5 	& 93.8 \\
$4<l \leqslant 8$& 509 	& 61.4	& 44.1	& 51.3	& 72.3 	& 79.7 	& 75.8 \\
$  > 8          $& 323 	& 30.3	& 15.0	& 20.0	& 74.0 	& 64.4 	& 68.9 \\
\hline
\end{tabular}
\caption{Comparison of Inside/Outside and Open/Close on SV patterns of
varying lengths.}
\label{CompareLength}
\end{center}
\end{table*}
\begin{table*}[!htb]
\vspace*{0.2in}
\begin{center}
\begin{tabular}{|l|c|c|c|c|}
\hline
Method	& Recall & Precision & $F_{\beta = 1}$ & Accuracy \\
\hline\hline
Inside/Outside           & 90.5 	& 90.4 	& 90.4 	& 97.0    \\
Inside/Outside + lexical & 92.5 	& 92.2 	& 92.4 	& 97.6    \\
Open/Close    		 & 90.9 	& 90.3 	& 90.6 	& O: 97.4, C: 97.8 \\
Open/Close + lexical   	 & 93.1 	& 92.4 	& 92.8 	& O: 98.1, C: 98.2 \\
\hline
Ramshaw \& Marcus        & 90.7 	& 90.5 	& 90.6 	& 97.0    \\
Ramshaw \& Marcus + lexical	& 92.3 	& 91.8 	& 92.0 	& 97.4    \\
Argamon {\em et al}.     & 91.6 	& 91.6 	& 91.6 	& N/A     \\ \hline
\end{tabular}
\caption{Comparison of Results for NP.  In the accuracy column, O indicates
the accuracy of the Open predictor and C indicates the accuracy of the Close
predictor.}
\label{CompNPResults}
\end{center}
\end{table*}

\begin{table*}[!htb]
\vspace*{0.2in}
\begin{center}
\begin{tabular}{|l|c|c|c|c|}
\hline
Method	& Recall & Precision & $F_{\beta = 1}$ & Accuracy \\
\hline\hline
Open/Close   			& 88.3 	& 87.9 	& 88.1 	& O: 98.6, C: 99.4 \\
Open/Close + lexical		& 91.9 	& 92.2 	& 92.0 	& O: 99.2, C: 99.4 \\
\hline
Argamon {\em et al.}           	& 84.5 	& 88.6 	& 86.5 	& N/A     \\ \hline
\end{tabular}
\caption{Comparison of Results for SV.  In the accuracy column, O indicates
the accuracy of the Open predictor and C indicates the accuracy of the Close
predictor.}
\label{CompSVResults}
\end{center}
\end{table*}

\section{Conclusion}
\label{sec:conclusion}

We have presented a SNoW based learning approach to shallow parsing tasks. The
learning approach suggests to identify a syntactic patterns is performed by
writing a simple program in which several instantiations of SNoW learning
units are chained and combined to produce a coherent inference.  Two
instantiations of this approach have been described and shown to perform very
well on NP and SV phrase detection.  In addition to exhibiting good results on
shallow parsing tasks, we have made some observations on the sensitivity of
modeling the task.
We believe that the paradigm described here, as well as the basic learning
system, can be used in this way in many problems that are of interest to the
NLP community.

\section*{Acknowledgments}
We would like to thank Yuval Krymolowski and the reviewers for their helpful
comments on the paper.  We also thank Mayur Khandelwal for the suggestion to
model the combinator as a graph problem.

\bibliographystyle{acl}
\bibliography{nlp,learn}
\end{document}